\documentclass[conference]{IEEEtran}
\let\labelindent\relax

\usepackage{times}
\usepackage[numbers]{natbib}
\usepackage{multicol}
\usepackage[bookmarks=true]{hyperref}
\usepackage{comment}
\usepackage{graphicx}
\usepackage{hyperref}
\usepackage{lettrine} 
\usepackage{makecell}
\usepackage{xcolor}
\usepackage{subfigure}
\usepackage{booktabs}
\usepackage{amsmath} 
\usepackage{amssymb} 
\usepackage{cleveref}
\usepackage{enumitem}

\newcommand{\eg}{\textit{e}.\textit{g}., }
\crefname{section}{Sec.}{Section.}
\Crefname{section}{Sec.}{Section.}
\crefname{figure}{Fig.}{Figure.}
\Crefname{figure}{Fig.}{Figure.}
\crefname{table}{Table.}{Table.}
\Crefname{table}{Table.}{Table.}
\crefname{equation}{}{}
\Crefname{equation}{Equation}{Equations}

\title{\LARGE \bf
Robustifying Long-term Human-Robot Collaboration \\ 
through a Multimodal and Hierarchical Framework
}

\author{\authorblockN{Peiqi Yu\authorrefmark{1}\authorrefmark{2}
Abulikemu Abuduweili\authorrefmark{1}\authorrefmark{2},
 Ruixuan Liu\authorrefmark{1}, and 
Changliu Liu\authorrefmark{1}}
\authorblockA{\authorrefmark{1}Robotics Institute, Carnegie Mellon University\\ Email: \{peiqiy, abulikea, ruixuanl, cliu6\}@andrew.cmu.edu}
}

\begin{document}

\maketitle

\footnotetext{These authors contributed equally to this work.}

 \begin{abstract}
Long-term Human-Robot Collaboration (HRC) is crucial for enabling flexible manufacturing systems and integrating companion robots into daily human environments over extended periods. This paper identifies several key challenges for such collaborations, such as accurate recognition of human plan, robustness to disturbances, operational efficiency, adaptability to diverse user behaviors, and sustained human satisfaction.
To address these challenges, we model the long-term HRC task through a hierarchical task graph and presents a novel multimodal and hierarchical framework to enable robots to better assist humans to advance on the task graph. In particular, the proposed multimodal framework integrates visual observations with speech commands to facilitate intuitive and flexible human-robot interactions. Additionally, our hierarchical designs for both human pose detection and plan prediction allow better understanding of human behaviors and significantly enhance system accuracy, robustness and flexibility. Moreover, an online adaptation mechanism enables real-time adjustment to diverse user behaviors. 
We deploy the proposed framework to KINOVA GEN3 robot and conduct extensive user studies on real-world long-term HRC assembly scenarios. Experimental results show that our approaches reduce task completion time by 15.9\%, achieves an average task success rate of 91.8\% and an overall user satisfaction score of 84\% in long-term HRC tasks, showcasing its applicability in enhancing real-world long-term HRC.
\end{abstract}

\IEEEpeerreviewmaketitle

\section{INTRODUCTION\label{sec:intro_challenge}}
Human-robot collaboration (HRC) has become a critical area of research as robots emerge in human-centric environments. 
Significant progress has been made in short-term HRC \cite{cheng2020towards, abuduweili2019adaptable, taskagnostic}, which addresses brief and localized interactions on single subtasks.
On the other hand, long-term HRC studies extended and hierarchical collaboration involving multiple interdependent subtasks.
For instance, in industrial settings, long-term HRC can significantly enhance productivity and adaptability in complicated manufacturing processes \cite{vysocky2016human}. 
Similarly in home environments, robots acting as companions or elderly care agents can notably improve quality of life through long-term HRC \cite{amirabdollahian2013assistive}. 
Despite its importance in many applications, long-term HRC remains under-explored \cite{amirova202110}.

Achieving effective and robust long-term HRC presents several challenges:  1) robots must accurately interpret human actions and predict human plans in complex, extended tasks; 2) systems must remain robust against environmental noise and unexpected disturbances; 3) interactions should be efficient and optimized in completion time for varying contexts; 4) systems should adapt to diverse user behaviors and preferences; 5) systems should minimize human workload, ensure ease of use, and enhance satisfaction throughout the prolonged collaboration process.
\begin{figure*}[htbp]
      \centering
      \includegraphics[width=1.02\linewidth]{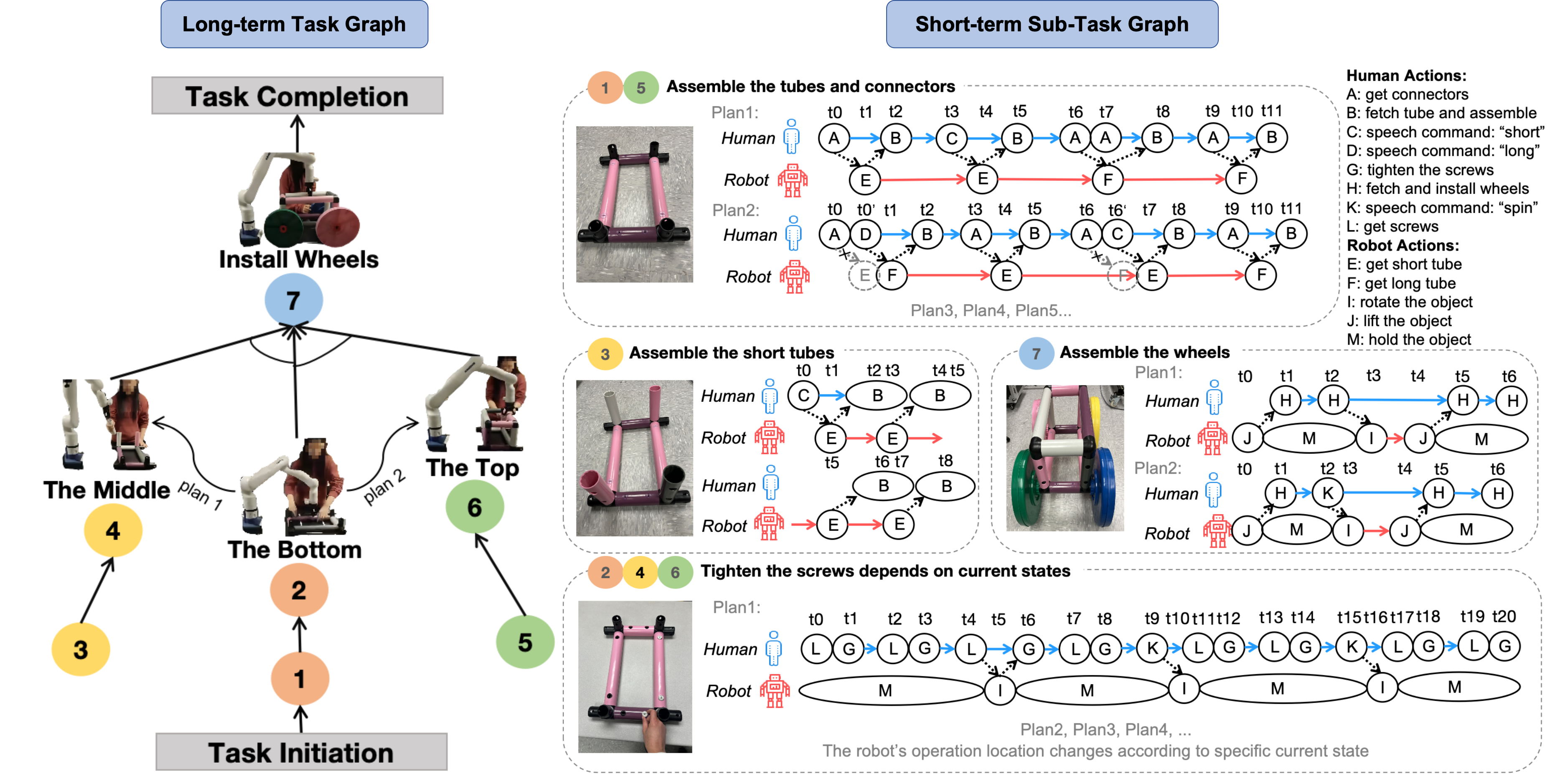}
      \caption{
      Overview of the hierarchical task graph for the long-term toy car assembly HRC task. Each node in the long-term graph (left) corresponds to a short-term subtask (right), where the human and robot collaborate through sequential actions. The task consists of four stages and each stage is completed through a series of task nodes (\eg must complete both Node 1 and Node 2 to complete The Bottom Stage ). The task starts at Node 1 and finishes at Node 7, with the directed arrows in the long-term task graph representing the temporal sequence of task execution (\eg must complete Node 1 before performing Node 2). Both high-level task planning and short-term execution involve uncertainties.
      In the subtask graph, dashed lines indicate collaborative actions, while solid lines represent actions performed by a single agent. Arrows denote task progression. Arrows with a cross indicate robot motions that were initially planned but were not executed, as the robot adapted to human actions and switched to an alternative motion. For clarity, certain actions are grouped or omitted without affecting the task’s overall structure.
      }
      \label{fig:overview-graph}
      \vspace{-10pt}
\end{figure*}

Addressing these challenges is essential for deploying long-term HRC in real-world applications. 
Although existing research tackles some aspects of these challenges, there remains a critical need for integrated solutions that address several challenges simultaneously. 
For instance, studies on adaptable HRC \cite{cheng2020towards,abuduweili2019adaptable,9281312} emphasize adaptable human motion prediction but often overlook robustness to disturbances or seamless integration required for effective HRC. 
By seamless integration, we refer to the robot’s ability to continuously interpret multimodal human inputs, adapt its behavior in real time, and transition smoothly between different subtasks without requiring frequent recalibration or explicit interventions \cite{papanastasiou2019towards}.
Similarly, multimodal approaches to human plan understanding \cite{liu2018towards, wang2024multimodal} rely heavily on wearable devices, limiting their generalizability across different interaction contexts. Furthermore, many studies lack unified frameworks and often fail to demonstrate their applicability in real-world long-term HRC tasks \cite{matheson2019human}.

In this work, we focus on enhancing long-term HRC characterized by extended, sequential interactions. 
We model long-term HRC tasks using a hierarchical task graph, assumed to be known to both human and robot participants. 
\cref{fig:overview-graph} illustrates an example of the task graph, consisting of a long-term task graph (left) and short-term subtask graphs (right). 
The long-term task graph structures the overall workflow, beginning at Node 1 with multiple pathways before converging at Node 7 (Assemble the Wheels) for task completion. The nodes are grouped into four stages. The short-term subtask graphs illustrate detailed human-robot interactions within each node, where sequential and collaborative actions are depicted. 

The goal for the human and the robot is to progress from the initial node to the completion node of the task graph through a series of interactions. 
In this paper, we define a human plan as the route of nodes that the human expects to advance in order to complete the task. 
We divide the human plan into two levels: high-level plan (node sequences in the long-term graph) and low-level plan (node sequences in the short-term subtask graph). 
Given the long sequences of these interactions, long-term HRC is prone to compounding errors due to uncertainties in both high-level and low-level plans (\eg the human can either go to node 3 or 5 from node 2 in the long-term task graph and may also have different low-level action plans in the sub-task graph). 
Therefore, precise and robust human plan predictions, and adaptability to individual users, are essential for effective long-term HRC.

To address these challenges, we propose a multimodal and hierarchical framework to enhance long-term HRC. The multimodality integrates visual and auditory signals, enabling accurate, reliable, and flexible predictions over dynamic and complex tasks. The hierarchical structure is reflected in two key components: hierarchical human pose detection and hierarchical plan prediction. In the plan prediction module, high-level human plans are first predicted based on the long-term task graph, and then refined through low-level plan predictions based on the subtask graph, improving flexibility to diverse user plans and enhancing system efficiency and task success rates. Hierarchical human pose detection further mitigates the effects of environmental noise and disturbances. Additionally, an adaptive action prediction module tailors the system to individual users. We apply the proposed multimodal hierarchical framework to a real-world human-robot co-assembly task. Experiment results demonstrate that our framework achieves precise, robust, flexible, and efficient collaboration, effectively addressing the critical challenges of long-term HRC.

In summary,  our main contributions are as follows:
\begin{enumerate}[leftmargin=12pt,itemsep=1pt, topsep=0pt]
\item We identify key challenges in long-term HRC tasks and propose a multimodal hierarchical framework that can effectively and robustly address these challenges.
\item We present a structured task graph for modeling long-term HRC tasks and deploy it to a real-world assembly scenario. 
Our proposed framework and task graph demonstrate successful HRC in the long-horizon assembly task.
\item We validate the framework through extensive physical experiments and user studies. The results show significant improvements in the robustness, time efficiency, and flexibility of HRC. Furthermore, user feedback highlights a strong preference for our framework, emphasizing its usability and practical applicability.
\end{enumerate}

\section{RELATED WORKS}
\textbf{Long-term Human-Robot Collaboration.}
Previous research on HRC has primarily focused on short-term tasks with long-horizon action sequences~\cite{pirk2020modeling,mishra2023generative}, rather than fully addressing complex long-term tasks. Although these studies explore important aspects of task planning in extended action sequences \cite{ng2023long,gupta2019relay}, there is limited work specifically targeting long-term HRC that achieves comprehensive and goal-oriented tasks. Pirk et al.~\cite{pirk2020modeling} successfully recognized long-term human subassembly actions, but did not demonstrate HRC in real-world scenarios. In this work, we implement our framework on a long-term real-world human-robot co-assembly task.

\textbf{Multimodal Human-Robot Collaboration.}
The interaction methods of the multimodal HRC framework can be summarized into four categories: vision, auditory and/or language, physiological sensing (human-centered sensing), and haptics (robot sensing) \cite{wang2024multimodal}. 
Physiological sensing \cite{fan2022vision,wang2024multimodal} typically requires extra wearable devices to detect bioelectrical signals, which often reduces generalizability across different interaction contexts and increases the operational complexity.  
Vision-language fusion-based multimodal human-robot collaboration frameworks have proven effective in tasks such as navigation and social interaction \cite{hong2020multimodal,chen2021history}. 
Liu et al. \cite{liu2018towards} demonstrated the effectiveness of model fusion of vision and auditory modalities in assembly tasks, although their approach has not been tested on real robots. 
Maurtua et al. \cite{maurtua2017natural}  relied on human gestures for vision perception and auditory input to conduct human-robot collaboration in industrial environments.
In this work, we fuse vision and auditory modalities, requiring only a camera and a microphone, in assembly tasks with potential applications for industrial and home service robots.

\textbf{Human Pose Detection and Physical Action Prediction.} 
High-fidelity and robust human behavior prediction is a key component of achieving safe HRC \cite{abuduweili2021robust,liu2023proactive}. Several approaches utilize human detection models, such as OpenPose \cite{cao2017realtime}, to identify keypoints on human body, which are then used to predict human behaviors including intentions and/or trajectories \cite{abuduweili2019adaptable,9281312}. 
Reducing disturbances in human pose detection has been extensively studied using signal filtering techniques \cite{welch2021kalman,abuduweili2020robust}. 
Traditional filtering methods primarily mitigate temporal disturbances \cite{sarafianos20163d}. In contrast, we focus on spatial aspects and propose a novel hierarchical framework to reduce disturbances in HRC. 
Additionally, we implemented an online model adaptation algorithm to adapt the prediction models to accommodate different users' behaviors \cite{abuduweili2023online}.

\section{PROBLEM FORMULATION}
\textbf{Long-Term HRC Task.}
This paper focuses on assistive collaboration in the human-robot supervisor-subordinate paradigm \cite{jarrasse2014slaves}. This paradigm leverages human strengths while enabling robots to reduce the physical and cognitive demands on humans by handling repetitive, heavy, or precision-intensive tasks. This approach is particularly well-suited for human-robot collaboration in manufacturing settings, where efficiency, workload distribution, and adaptability are critical.

We consider a scenario where a human $A_H$ collaborates with a robot $A_R$, to achieve a set of predefined goals within a long-term task $\mathcal{G}$. The task $\mathcal{G}$ is represented as a hierarchical directional And-Or graph \cite{andorgraph}, with the ultimate goal $g$ defined as reaching the end node of the graph (see \cref{fig:overview-graph} left). The nodes $\mathcal{N}$ of $\mathcal{G}$ correspond to subtasks, and the directed edges represent precedence relationships. We group several nodes into one stage, and each stage may have multiple successors, reflecting the possibility of alternative plans (\eg ''The Bottom" stage in \cref{fig:overview-graph} can lead to either ''The Middle" or ''The Top" stage).

Long-term HRC tasks pose significant challenges due to uncertainties and disturbances, particularly in human behavior. In this work, we focus on human-related uncertainties while assuming object positions are well-observed, as human behavior and actions present greater unpredictability. Key challenges include variability (including diverse preferences and styles) in human behavior, visual ambiguities caused by occlusions or background clutter, and uncertainty in human plans at both high and low levels of the task graph. Additionally, environmental disturbances and compounding errors over prolonged interactions exacerbate these challenges. To address these challenges, we aim to robustify long-term HRC by focusing on human behavior-related uncertainties and disturbances. 
   
\textbf{Sub-task Configuration}. Nodes $\mathcal{N}$ in the long-term task $\mathcal{G}$ corresponds to shorter-term HRC tasks, referred to here as subtasks.
As shown in the right part of \cref{fig:overview-graph}, each subtask $N_i \in \mathcal{N}$ is organized as a multi-agent Temporal Plan Graph (TPG), which represents both the precedence and collaborative dynamics between human and robot actions \cite{feng2024real}. A Temporal Plan Graph (TPG) \cite{honig2016multi} captures precedence relationships within a HRC and preserves them during execution. Each vertex in $N_i$ represents an action performed by a human or a robot. 

\textbf{Definition of Hierarchical Human Plan}.
The human plan refers to the route of nodes that the human is expecting to advance on to complete the task. We divide human plan $x_H$ into two levels, where the low-level plan is a sequence of nodes in a short-term sub-task graph and the high-level plan is a sequence of nodes in a long-term task graph. \Cref{fig:overview-graph} illustrates that the user may have different high-level plans as well as different low-level plans within the same high-level plan. To complete the long-term HRC task, the robot must predict both human plans precisely and perform corresponding responses.

\textbf{The Objective Function.}
Let $E$ denote the environment state in the HRC task, including human actions, object positions, and surrounding context. $E$ is derived from sensor observations, which may include noise or ambiguities in human behavior.  Let $P$ denote the task progress, which tracks the completion of subtasks (or the current node within the task graph $\mathcal{G}$) by incorporating both the sequence of human-robot actions executed and the overall task graph. As the collaboration progresses, $P$ is dynamically updated. At timestep $t$, the human's plan $x_H^t$ is influenced by the current state of the environment $E^t$ and task progress $P^t$.   
The robot predicts human plans $\hat{x}_H^t$ using its learnable model  $\hat{f}_H$:
\begin{align}
   \hat{x}_H^t = \hat{f}_{H} (E^t, P^t), \label{eq:human_intent_pred}
\end{align}
The predicted plans guide the robot in generating a motion sequence $q = [q^1,\cdots,q^n]$.  The objective for the robot is to find an efficient and robust motion sequence $q$ under safety and dynamics constraints. Here, $n$ represents the planning horizon, which is determined based on the specific task, the robot’s capabilities, and the reliability of the predictions. For example, 
$n$ may correspond to planning the robot's motion for the next 2 seconds based on the current predicted human plans.
Let $J_g$ represent the task-specific cost of achieving the goal $g$, such as the tracking cost.  To reflect time efficiency, we include the task completion time $T_g$, weighted by $\lambda$, in the objective function. Thus, the robot's objective in HRC is formulated as:
\begin{align}
    & \min_{q= [q^1,\cdots,q^n]} ~ J_g(q, g) + \lambda T_g(q, g) \label{eq:obj} \\
    & \text{s.t.} ~ \forall t \in [1,n], ~ q^{t} = M_R(\hat{x}_H^t)
\end{align}
where $M_R(\cdot)$ represents the robot motion planning function. 
This objective is influenced by the environment state $E^t$,  task progress $P^t$, and predicted human plan $\hat{x}_H^t$. While $E^t$ and $P^t$ are relatively deterministic, the uncertainty in human plans and variability in human behavior significantly impact the generation of optimal robot motions. Thus, robust and accurate prediction of human plans, along with mitigating the effects of noise on human behavior, is crucial for achieving optimal motion generation. This, in turn, is essential for efficient and robust HRC.

\section{PROPOSED FRAMEWORK}
\begin{figure*}[thpb]
      \centering
      \includegraphics[width=\linewidth]{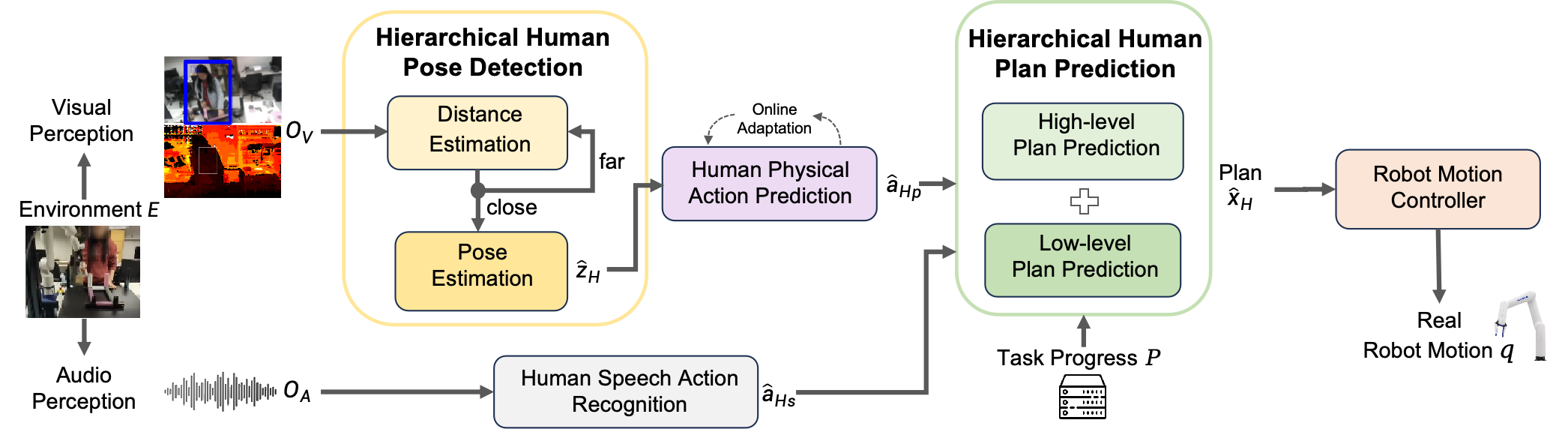}
      \caption{The architecture of the proposed HRC framework.}
      \label{fig:architecture}
      \vspace{-10pt}
\end{figure*}

To address the challenges of long-term HRC tasks as discussed in \cref{sec:intro_challenge}, we propose a hierarchical and multimodal HRC framework, as illustrated in \cref{fig:architecture}. 
The framework includes two perception modules, four task planning modules, and a robot motion controller.
\begin{itemize}[leftmargin=12pt,itemsep=1pt,  topsep=0pt]
    \item  \textbf{Perception.} \textit{Visual Perception Module} captures visual information from the environment and humans using  RGBD cameras.  \textit{Audio Perception Module} processes audio information through microphones.
    \item  \textbf{Planning.} \textit{Human Pose Detection Module} identifies human poses and extracts key point trajectories. Its hierarchy enhances detection robustness by mitigating disturbances in complex, multi-human environments. \textit{Human Physical Action Prediction Module} predicted future human pose trajectories and interpret the semantic action label of the current human pose. \textit{Human Speech Action Recognition Module} recognizes human speech commands and outputs human speech action labels. \textit{Human Plan Prediction Module} integrates physical and speech actions and task progress to predict human plans. Its hierarchical design improves prediction accuracy and enhances system robustness in responding to human actions \cite{cheng2020towards}. 
    \item \textbf{Control.} \textit{Controller Module} synthesizes predicted human plans and advances on the task graph by generating executable robot motions.
\end{itemize}

\subsection{Multimodal Perception}\label{subsec:multimodality}
In human-human interactions, visual observations alone often fail to fully capture human plans, necessitating the integration of verbal communication for comprehensive context.  Extending this principle to long-term HRC tasks, we propose a multimodal framework (depicted in \cref{fig:architecture}) that combines visual and audio observations for more comprehensive perception. As the same human physical action may indicate different plans depending on the context, we incorporate speech actions to enable humans to clarify and refine robot-predicted human plans derived from visual interpretations. The multimodal approach offers two key advantages: 1) By integrating visual and audio signals, the robot achieves a comprehensive perception of human plans, allowing for more complex and long-term collaborations. 2) The inclusion of speech enables more natural and flexible interactions, making the system more user-friendly and accessible.

Let $O_V$ denote the robot's visual observation from the environment $E$, captured by the RGBD camera; and $O_A$ denote the robot's audio observation, captured by the microphone. It can be easily deducted that multual information $I(g; O_V, O_A) \ge \max \{  I(g; O_{V}) ,  I(g; O_{A})$\}. The proof of this inequality is provided in the appendix. This inequality demonstrates that the multimodal framework yields greater mutual information than single-modal approaches, significantly enhancing the robot's ability to manage complex and long-term collaborative tasks.

\subsection{Hierarchical Human Pose Detection}\label{subsec:hierarchical-human-detection}
Human poses are fundamental to understanding human physical actions, which in turn are critical for predicting human plans. However, visual observations can often be noisy, leading to errors in pose detection and subsequent plan prediction. This noise can arise from various sources, such as the presence of multiple humans in the scene or cluttered backgrounds. To mitigate these disturbances, we propose a hierarchical pose detection module. 

Our module operates in two steps, as illustrated in the yellow box in \cref{fig:architecture}. First, we detect all humans in the scene and estimate their distances from the robot through an RGB-D camera. If no human is detected within a predefined range (\eg 2 meters), the detection cycle repeats, and downstream HRC tasks are temporarily paused. This ensures the robot focuses on humans likely to interact closely based on proximity. Once a human is detected within the effective range, we crop on the closest human. This cropped image is then processed by the pose estimation model to extract its key points. This design significantly reduces keypoint deviation, thus help to make pose detection more robust, as shown in \cref{fig:concatenated_mask_no_image}. 

This hierarchical structure offers two key advantages: 1) By activating pose estimation and planning modules only when a human is detected within range, the system conserves computational and energy resources. 2) The hierarchical filtering reduces interference from non-target humans and background noise, improving pose detection accuracy. The following equation represents the operation of our proposed hierarchical pose detection model $ \hat{f}_{\rm detection}$, which estimates the human pose  $\hat{z}_H^t$ from visual observations: $\hat{z}_H^t = \hat{f}_{\rm detection} (O_V^t) \label{eq:human_detect}$.

\subsection{Human Physical Action Prediction}\label{subsec:human_physical_action_prediction}
Rather than predicting human actions solely based on current pose trajectories, our approach first predicts future trajectories and then infers actions using both the current and predicted trajectories. This can enhance system performance by ensuring that the action prediction is informed by anticipated future movements \cite{abuduweili2019adaptable}.
We jointly train trajectory and action prediction models using supervised learning, utilizing collected offline data. The trajectory prediction model is optimized using regression loss (L2 loss), while the action prediction model employs classification loss (cross-entropy loss). Once trained, the action prediction model $ \hat{f}_{\text{action}}$ predict human actions $\hat{a}_{H_p}^t$ and its output confidence based on the current human pose: $\hat{a}_{H_p}^t, \hat{\epsilon}_{H_p}^t = \hat{f}_{\text{action}}(\hat{z}_H^t).$
\begin{table}[ht]
  \centering
  \begin{tabular}{p{5.2cm}<{\centering} p{2.1cm}<{\centering}}
    \hline
    \textbf{Error Type} & \textbf{Sensitivity score} \\
    \hline
    ``No Action" predicted as other actions & high  \\
    One action predicted as another action & medium \\
    One action predicted as "No Action" & low \\
    \hline
  \end{tabular}
  \caption{Sensitivity score of different misclassification errors.}
  \label{tab:Sensitivity_score}
  \vspace{-10pt}
\end{table}

Additionally, we assign different sensitivity scores to various types of misclassification errors and optimize the prediction model to minimize the most critical errors. The sensitivity scores assigned to each type of error are shown in \cref{tab:Sensitivity_score}. We propose two restriction strategies to mitigate higher-sensitive errors: confidence and boundary restriction, the details and experimental results are shown in \cref{subsubsec:exp_h2}.

To enhance model generalization to different human behaviors, we implement an online adaptation method \cite{abuduweili2021robust} during deployment, enabling the robot to refine predictions in real-time by learning each user's behavior \cite{cheng2020towards}. Our joint training framework supports dynamic adaptation by optimizing the trajectory prediction model online using observed trajectory feedback. As action predictions depend on the trajectory module, updates to the trajectory model directly improve action prediction accuracy \cite{abuduweili2020robust}. The online adaptation approach implements the method reported in \cite{abuduweili2023online}. This technique significantly improves prediction performance, ensuring that the robot’s responses accurately align with each user’s behaviors and preferences.

\subsection{Human Speech Action Recognition}
\label{subsec:speech_action}
For the audio modality $O_A$, the robot recognizes speech action $\hat{a}_{H_s}^t$ along with its confidence using a pretrained speech recognition model $\hat{f}_{\rm speech}$: $\hat{a}_{H_s}^t, \hat\epsilon_{H_s}^t = \hat{f}_{\rm speech}(O_A)$.  In deployment, this model's parameters are fixed. To enhance the quality of speech recognition, a Voice Activity Detector (VAD) is utilized to filter the input. This VAD activates the speech recognition module only when significant audio signal energy, not noise, is detected.

\subsection{Hierarchical Human Plan Prediction}\label{subsec:hierarchical-plan-prediction}

After obtaining human physical action $\hat{a}_{H_p}^t$ and human speech action $\hat{a}_{H_s}^t$, the robot determine the final human action $\hat{a}_{H}^t$ based on priority (speech command ``stop" has the highest priority. In other cases, priority is determined by the action with the higher confidence level). The robot then predicts human plan $\hat{x}_{H}^t$ based on the final human action $\hat{a}_{H}^t$. 
We divide this process into two hierarchical levels. Low-level plan prediction focuses on short-term sub-task nodes, such as “get screws”, which dictate immediate action sequences. High-level plan prediction captures broader human intent, corresponding to long-term task sequences. 

The robot identifies the reference node sequence $R^\ast$ within the task graph that most closely aligns with the recorded task progress $P^t$. This alignment is formulated as:
\begin{align}
    R^\ast = \arg \min_{R \in  \mathcal{G}} d(P^t, R),
\end{align}
where $d$ represents the distance between the observed task progress $P^t$ and each possible reference sequence $R$ within the task graph. The distance metric $d$ is calculated using the dynamic time warping (DTW) algorithm \cite{tormene2009matching}. The result, $R^\ast$, represents the best-matched reference node sequence. Note that $R^\ast$ may not be unique, as multiple node sequences may present equal distances, allowing for different pathways to the same goal. 
For instance, in the long-term task graph \cref{fig:overview-graph}, the assembly process may start from node 1, 3, or 5, indicating different initial conditions. Suppose the observed task sequence suggests that the human is assembling the short tubes (Plan 3). The robot aligns the observed task progress $P^t$
 with the closest reference sequence $R^\ast$ within the plan graph.

Following this alignment, the human plan prediction is obtained by refining the predicted human action based on the most probable action class within  $R^\ast$. This adjustment is formulated as:
\begin{align}
    \hat{x}_{H}^t = R^\ast(\hat{a}_{H}^t),
\end{align}
where $\hat{x}_{H}^t$ is the predicted human plan, aligning $\hat{a}_{H}^t$ with the most likely action from $R^\ast$. 
For example, as shown in the sub-task graph of \cref{fig:overview-graph}, in Plan 1 of assembling tubes and connectors, if the human has already performed action A (fetching connectors), the most probable next action in $R^\ast$ is Action B (fetching tubes and assembling them). The robot adjusts its prediction accordingly, ensuring its motion planning remains synchronized with human execution.

By hierarchically structuring  human plan prediction into two levels and aligning the robot's predictions with the task graph and task progress,  this method improves the accuracy and flexibility of human plan prediction.

\subsection{Robot Motion Controller}
\label{subsec:motion_controller}
Upon acquiring the human plan $\hat{x}_H^t$, the robot's motion planner generates the robot motions according to human plan and current task progress:
\begin{align}
   q^{t:t+m} &= M_R (\hat{x}_H^t)
   \label{eq:robot_motion}
\end{align}
where $ q^{t:t+m} $ represent the robot motion sequences and $M_R$ represents the motion controller. The motion controller operates as a feedback controller with an update frequency of approximately 30 Hz. 
In accordance with the optimization objective, accurate prediction of $\hat{x}_H^t$ is crucial for optimal motion generation within the task graph. Errors in $\hat{x}_H^t$ disrupt robot motions, reducing efficiency and robustness in HRC.

\section{EXPERIMENTS}
\subsection{Toy Car Assembly Task} \label{sec:ex_setting}
\begin{figure}[thpb]
      \centering
      \includegraphics[width=0.7\linewidth]{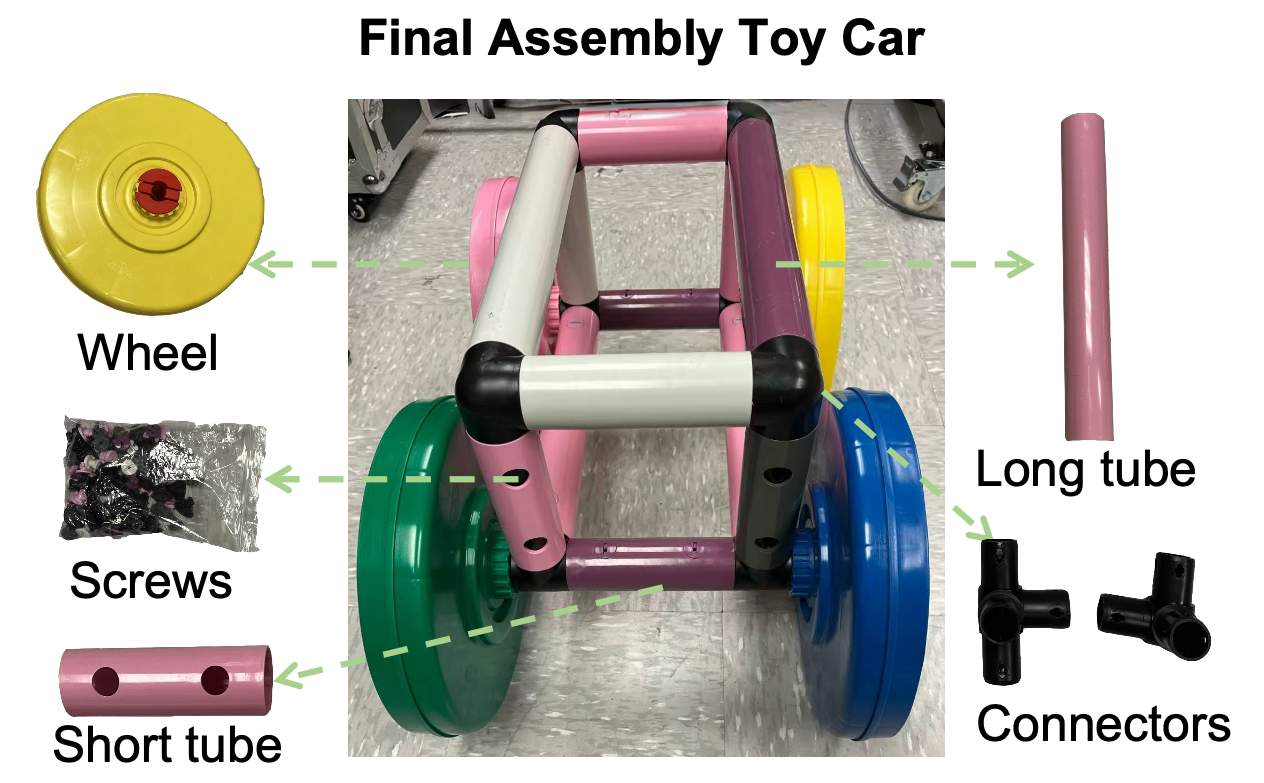}
      \caption{Final Assembly Task}
      \label{fig:final_assembly}
      \vspace{-5pt}
\end{figure}

We evaluate the effectiveness of our proposed framework through a long-term HRC task,  specifically the \textbf{toy car assembly} task, as shown in \cref{fig:final_assembly}. This assembly process incorporates various components, including both short and long tubes, connectors, screws and wheels.  This task is divided into four stages: assembling the bottom, the middle, the top of the car, and installing the wheels, as shown in \cref{fig:overview-graph}. For simplicity, we will refer to the four stages as \textit{The Bottom}, \textit{The Middle}, \textit{The Top} and \textit{Install Wheels}. The assembly order for \textit{The Middle} and \textit{The Top} is flexible, allowing variations in the sequence order. 

Each stage is designed to require collaborative efforts, in which the robot's role is to support the human collaborator by delivering necessary objects and performing specific assistive actions, such as rotating or lifting objects. The actions and motion sequences within these stages are non-deterministic, offering a range of valid action sequences that can successfully complete each sub-task. This flexibility means that the specific sequence of actions may vary according to individual human preferences. For instance, in \cref{fig:overview-graph}, completing Node 1 (or Node 5), which involves assembling tubes and connectors, can be achieved through various action sequences depending on the human's plan. Consider Plan 2 as an example: the human begins by getting the connectors (Node A). At this point, the robot predicts the human's plan, based on a default prediction model, that the human intends to "get the short tube" (Node E). However, the human instead verbalizes "long" (Node D),  correcting the robot's prediction, which then adjusts its action to get the long tube (Node F). After obtaining the correct component, the human proceeds to assemble the tube (Action B). Subsequently, the human gets the connectors again, and this time, the robot correctly predicts the human's intention to "get the short tube," aligning with the expected plan. Through dynamic adjustments to its assistance based on predicted human actions, the robot ensures seamless and flexible collaboration across various human plans, ultimately leading to the efficient completion of the overall task.
\begin{figure}[thpb]
      \centering
      \includegraphics[width=0.9\linewidth]{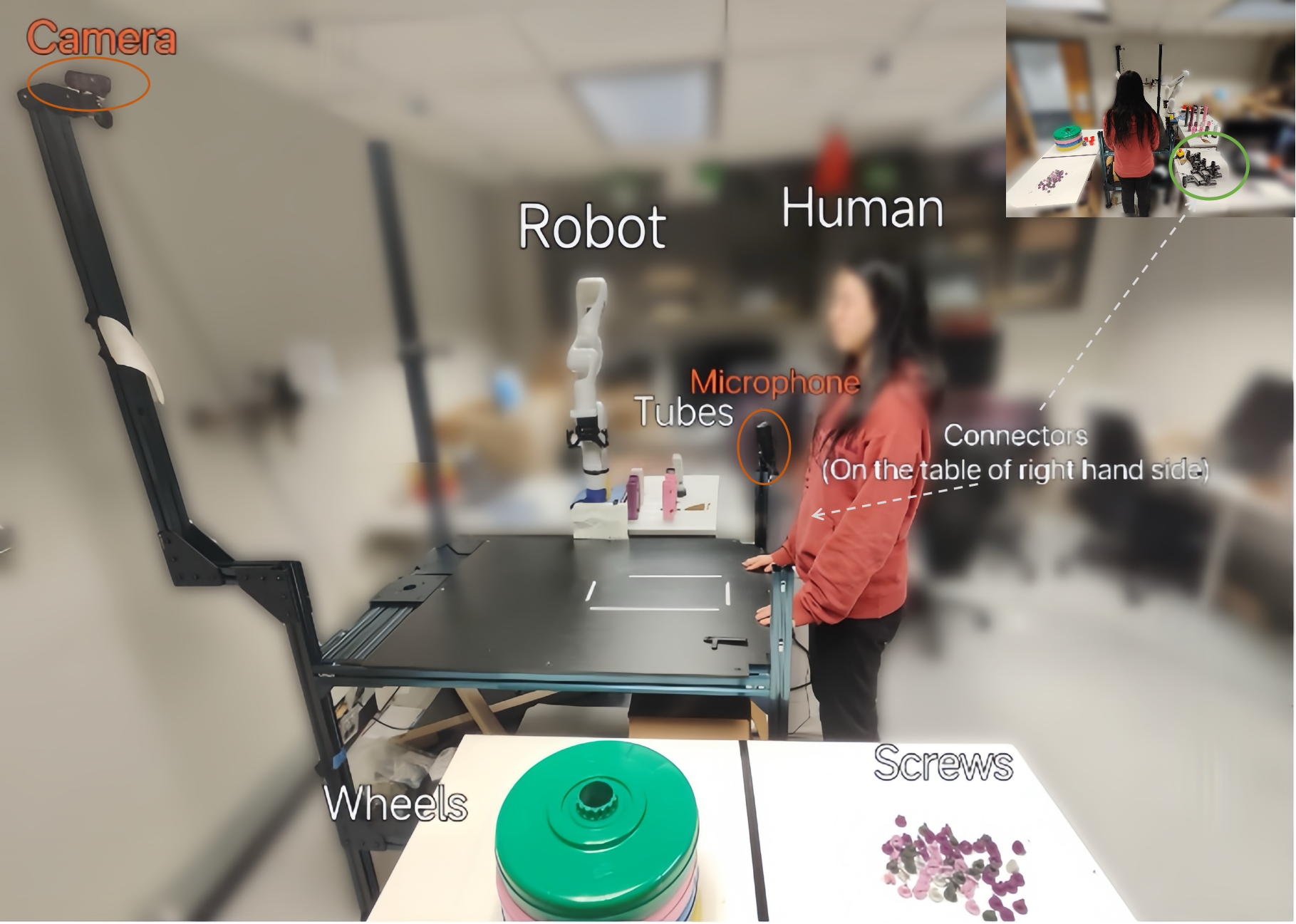}
      \caption{Environment Setting}
      \label{fig:environment_setting}
      \vspace{-10pt}
\end{figure}

The environment setup is shown in \cref{fig:environment_setting}, as the tools and objects needed for the assembly task are positioned across different locations. Connectors, screws, wheels, and screwdrivers are placed near the human to allow easy access. Conversely, the tubes are stored on a table that is closer to the robot arm and relatively farther from the human. The locations of all objects are fixed, and we assume that both the robot and the human know all the exact locations and there is no uncertainty in object perception. During the assembly process, the robot assists according to its understanding of human plans, current task progress and current environment. For instance, if the human plans to assemble tubes by reaching for connectors, the robot responds by delivering the appropriate tube—either short or long—based on the observed signals. When the human plans to retrieve screws, the robot aids by spinning the base-object to facilitate easier handling. Similarly, when the task involves attaching wheels, the robot assists by lifting the car's body to better position it for wheel installation.
The source code and demonstration video for the HRC toy car assembly task are available at { \url{https://github.com/intelligent-control-lab/Robust-Hierarchial-Multimodal-HRC}}.

\subsection{Experimental Design}
\subsubsection{Implementation Details}\label{subsubsec:implementation_details}
We test our Human-Robot Collaboration (HRC) framework using the KINOVA GEN3 robotic arm, equipped with an OAK-D Lite camera for video (capturing both RGB and depth maps) and a directional microphone for audio. Below, we detail the neural network models used across various components of our framework, noting that only the action prediction network is trained; all others utilize pretrained weights.
\begin{itemize}[leftmargin=12pt,itemsep=1pt,topsep=0pt]
    \item \textbf{Human Distance Estimation.} We first use a pretrained MobileNet model \cite{sandler2018mobilenetv2} to detect  ``human" from RGB data, applying a confidence threshold of 0.6. Then we estimate human distance from corresponding depth image.
    \item \textbf{Human Pose Estimation.} The pretrained BlazePose model \cite{bazarevsky2020blazepose} extracts 33 key points, of which we focus on 15 upper-body points for physical action prediction.
    \item \textbf{Human Physical Action Prediction.} We implement the DLinear model \cite{zeng2023transformers} for action and trajectory prediction, which decomposes input time series into trend and seasonal components. This model is trained using offline data to be specified in \cref{subsubsec:dataset}. We predict four human physical actions: ``get connectors"(Action1), ``get screws"(Action2), ``get wheels"(Action3), and No Action, where ``No Action" indicates instances where the human's action did not match any of other three actions.
    \item \textbf{Human Speech Action Recognition.} The DeepSpeech model \cite{amodei2016deep}, an open-source Speech-to-Text engine is utilized for recognizing and processing speech commands. We recognize five speech commands: ``short", ``long", ``spin", ``lift", and ``stop".
\end{itemize}

\subsubsection{Hypothesis} 
We evaluate the effectiveness of the proposed multimodal hierarchical HRC framework by testing five hypotheses:
\begin{itemize}[leftmargin=12pt,itemsep=1pt,topsep=0pt]
    \item H1: The framework increases task success rates in collaborative tasks.
    \item H2: The framework reduces disturbances in multi-human environments and increases action prediction accuracy.
    \item H3: The framework reduces task completion time and enhances the efficiency of HRC.
    \item H4: The framework adapts to varying human behaviors.
    \item H5: The framework allows users to be more flexible and improves user satisfaction in collaboration tasks.
\end{itemize}

\subsubsection{User Study}\label{subsubsec:user_study}
To evaluate the proposed framework and test the five hypotheses, we conducted a user study involving 10 participants. The study divides the toy car assembly task into three consecutive sub-tasks:
\begin{itemize}[leftmargin=12pt,itemsep=1pt,topsep=0pt]
    \item Task 1: Assemble tubes and connectors in \textit{The Bottom} Stage.
    \item Task 2: Based on the completion of \textit{The Bottom} Stage, complete \textit{The Middle} Stage and \textit{The Top} Stage according to the user plan.
    \item Task 3: Based on the completion of the previous three stages, complete \textit{Install Wheels} Stage.
\end{itemize}

We define ``Perception Modality = 0" (referred to ``PM" below) as vision-only perception, ``PM = 1" as audio-only perception and ``PM = 2" as multimodal perception, ``Plan Prediction = 0" (referred to as ``PP" below) as having no plan prediction and ``PP = 1" as having plan prediction. When ``PP = 0", the robot has a pre-defined plan sequence and only reacts after the human takes certain actions. When ``PP = 1", the robot hierarchically predicts human plans and reacts proactively. By defining these two variables, we have six groups of test conditions. Under each group, every user performs each task using any plan for four times. Thus, there will be 120 trials in each group and we collect 720 trials in total.

\subsubsection{Dataset}\label{subsubsec:dataset}
We collected an offline dataset to train and evaluate the action prediction module. The training data includes 3,003 trajectories from two users, while the test data consists of 2,088 trajectories from these two users and two additional users to evaluate robustness. Each sequence of trajectories contains 5 time steps.
The five hypotheses are evaluated using offline test data, user studies, or both.

\subsection{Results} \label{subsec:results}
\subsubsection{H1 - Task Success Rate} \label{subsubsec:exp_h1}
\Cref{tab:task_success_rate} compares the task success rates of six groups for each of the three sub-tasks outlined in the user study in \cref{subsubsec:user_study}.  The table summarizes results from 40 trials for each task and group. 
As shown in \cref{tab:task_success_rate}, the group with multimodal perception and hierarchical plan prediction (``PP = 1, PM = 2") achieves the highest task success rate (91.79\%). Notably, Task 2 demonstrates the most pronounced benefit of multimodal perception (``PM = 2"). In this task, the vision-only group (``PM = 0") struggles significantly when attempting to prioritize the \textit{Middle Stage}, resulting in a big decline of success rate (40.5\% on average compared to ``PM = 2"). Furthermore, integrating hierarchical plan prediction with multimodal perception (``PP = 1, PM = 2") increases success rate by 14.22\% compared to without plan prediction (``PP = 0, PM = 2"). These results demonstrate the synergistic advantages of combining hierarchical plan prediction and multimodal perception, particularly in handling complex tasks and user plans.

\begin{table}[ht]
\centering
\begin{tabular}{llccc}
\hline
\textbf{PP} & \textbf{PM} & \textbf{Task1 (\%)} & \textbf{Task2 (\%)} & \textbf{Task3 (\%)} \\
\hline
0 & 0 & $74.45 \pm 17.15$ & $48.34 \pm 6.21$ & $56.46 \pm 18.38$ \\
& 1 & $92.34 \pm 4.68$ & $82.32 \pm 19.18$ & $69.58 \pm 16.74$ \\
& 2 & $89.12 \pm 9.56$ & $83.21 \pm 12.21$ & $60.45 \pm 17.39$ \\
\hline
1 & 0 & $70.63 \pm 18.65$ & $50.12 \pm 8.26$ & $68.75 \pm 22.24$ \\
& 1 & $90.63 \pm 6.75$ & $81.25 \pm 19.98$ & $63.75 \pm 18.35$ \\
& 2 & $\textbf{93.75} \pm \textbf{9.32}$ & $\textbf{96.25} \pm \textbf{4.37}$ & $\textbf{85.38} \pm \textbf{13.58}$ \\
\hline
\end{tabular}
\caption{Comparison of task success rates.}
\label{tab:task_success_rate}
\vspace{-10pt}
\end{table}
   
\subsubsection{H2 - Disturbance Minimization}  \label{subsubsec:exp_h2}
\begin{figure}[thpb]
      \centering
      \includegraphics[width=\linewidth]{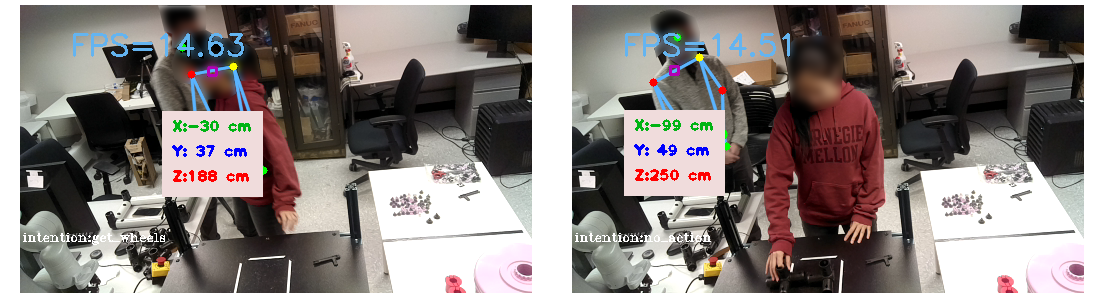}
      \caption{Misdetection in naive human pose detection model.} 
      \label{fig:concatenated_no_mask_image}
      \vspace{-5pt}
\end{figure}

\begin{figure}[thpb]
      \centering
      \includegraphics[width=\linewidth]{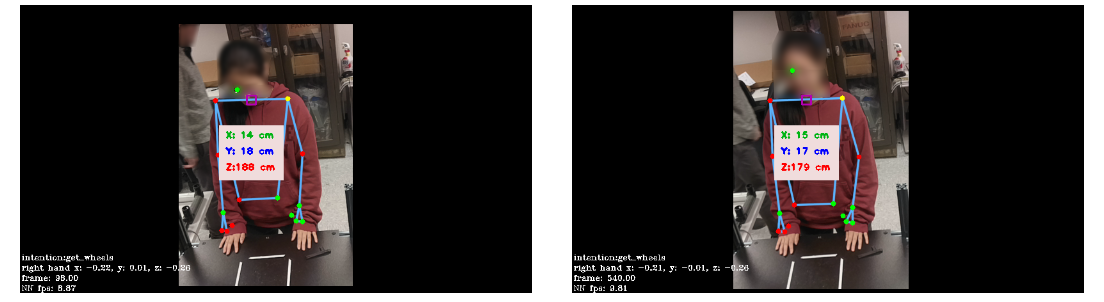}
      \caption{Correct detection using the hierarchical pose detection framework.}
      \label{fig:concatenated_mask_no_image}
      \vspace{-5pt}
\end{figure}

The presence of multiple humans in the same environment can introduce errors in human detection.
As discussed in \cref{subsec:hierarchical-human-detection}, we propose a hierarchical human detection module to address the challenges and disturbances caused by multiple humans. For comparison, sample results from a naive human pose detection model (without hierarchical detection) are shown in \cref{fig:concatenated_no_mask_image}.  As illustrated, the naive detection model can result in errors or misdetections when multiple humans are present. In contrast, the proposed hierarchical framework reduces these disturbances by prioritizing the human closest to and interacting with the robot, as demonstrated in \cref{fig:concatenated_mask_no_image}.

\begin{figure}[thpb]
      \centering
      \includegraphics[width=0.9\linewidth]{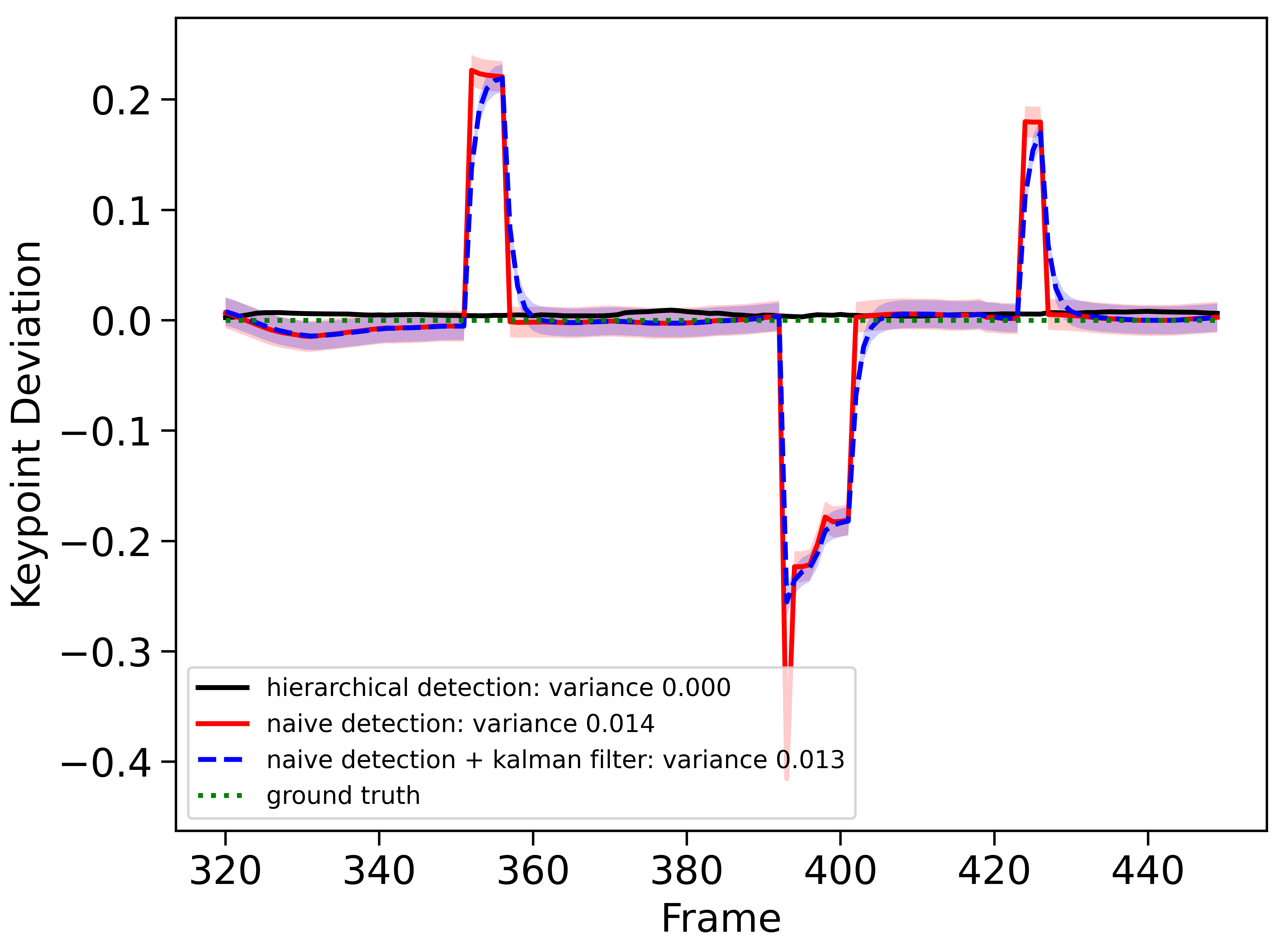}
      \caption{Keypoint deviation with and without hierarchical human detection.}
      \label{fig:mask_ablation_curve}
      \vspace{-10pt}
   \end{figure}

\textbf{Human Keypoints Detection.}    
We conducted the following case study to numerically evaluate the effectiveness of the hierarchical pose detection method on human keypoints detection. In this setup, one person stood close to the robot, serving as the human operator, while another person walked through the environment to introduce noise for the detection model. 
To evaluate the method's robustness, we measured the keypoint deviation over time.
\Cref{fig:mask_ablation_curve} shows the keypoint deviation for three methods: naive detection without hierarchical human detection, naive detection with Kalman filtering, and hierarchical human detection. The ground truth is represented by a constant zero line (green dotted line). As illustrated, the naive detection method, even when smoothed with Kalman filtering, results in sharp fluctuations in the detected keypoints, with a variance of 0.013. In contrast, the proposed hierarchical human detection method (black curve) provides robust and stable detection, closely aligning with the ground truth. These results support the hypothesis that the hierarchical pose detection method effectively minimizes disturbances in multi-human environments.

\textbf{Action prediction.} 
\Cref{tab:intention_comparison_hierarchical} compares the action prediction accuracy of the proposed human physical action prediction module against a naive baseline and four baselines using filtering methods in multi-human noisy environments. The methods are: (a) ``w/o hierar," which predicts actions using a naive human pose; (b) ``median/wiener/kalman/ema filter," which applies respective filters to the naive human pose before predicting actions; and (c) ``w/ hierar," which uses the hierarchical detection framework for action prediction. In the table, ``All" represents the overall average accuracy across all actions.
The results clearly show that hierarchical human detection significantly improves action prediction accuracy compared to other methods.

\begin{table}[htbp]
  \centering
  \begin{tabular}{p{1.2cm}<{\centering} p{0.8cm}<{\centering} p{0.7cm}<{\centering} p{0.7cm}<{\centering} p{0.7cm}<{\centering} p{0.7cm}<{\centering} p{0.8cm}<{\centering}}
    \hline
    & {w/o hierar} & {median filter} & {wiener filter} & {kalman filter} & {ema filter} & {w/ hierar} \\
    \hline
    No Action & 0.5824 & 0.4291 & 0.4713 & 0.2835 & 0.3295 & \textbf{0.9127} \\
    Action1 & 0.9810 & 0.8659 & 0.8644 & 0.5729 & 0.7930 & \textbf{0.9985} \\
    Action2 & 0.2196 & 0.6738 & 0.7234 & 0.8095 & 0.6211 & \textbf{0.9711} \\
    Action3 & 0.3835 & 0.6627 & 0.6406 & 0.7992 & 0.8996 & \textbf{0.9200} \\
    All & 0.5139 & 0.7001 & 0.7190 & 0.6811 & 0.6971 & \textbf{0.9617} \\
    \hline
  \end{tabular}
  \caption{Comparison of action prediction accuracy.}
  \label{tab:intention_comparison_hierarchical}
  \vspace{-15pt}
\end{table}

To minimize critical class-dependent misclassification errors—such as mistakenly classifying ``stay constant" as ``reach out to human"—we implement sensitivity-aware action prediction, which strategically "transfers" higher-sensitivity errors to low-sensitivity ones as defined in \cref{subsubsec:implementation_details}, improving overall prediction accuracy. 
We propose two restriction strategies and conduct experiments on offline data to demonstrate their effectiveness. 

The first strategy is confidence-based restriction, which assign different confidence scores to different actions. If the predicted probability for an action $a_1$ (other than "no-action") is lower than a confidence threshold (\textit{i.e} 0.6), we label the action as "no action" $a_0$  instead, even if the original probability of the action $a_1$  is higher than "no action".  This approach helps "transfer" a high-sensitivity error to a lower-sensitivity error.  

\begin{figure}[thpb]
      \centering
      \includegraphics[width=0.9\linewidth]{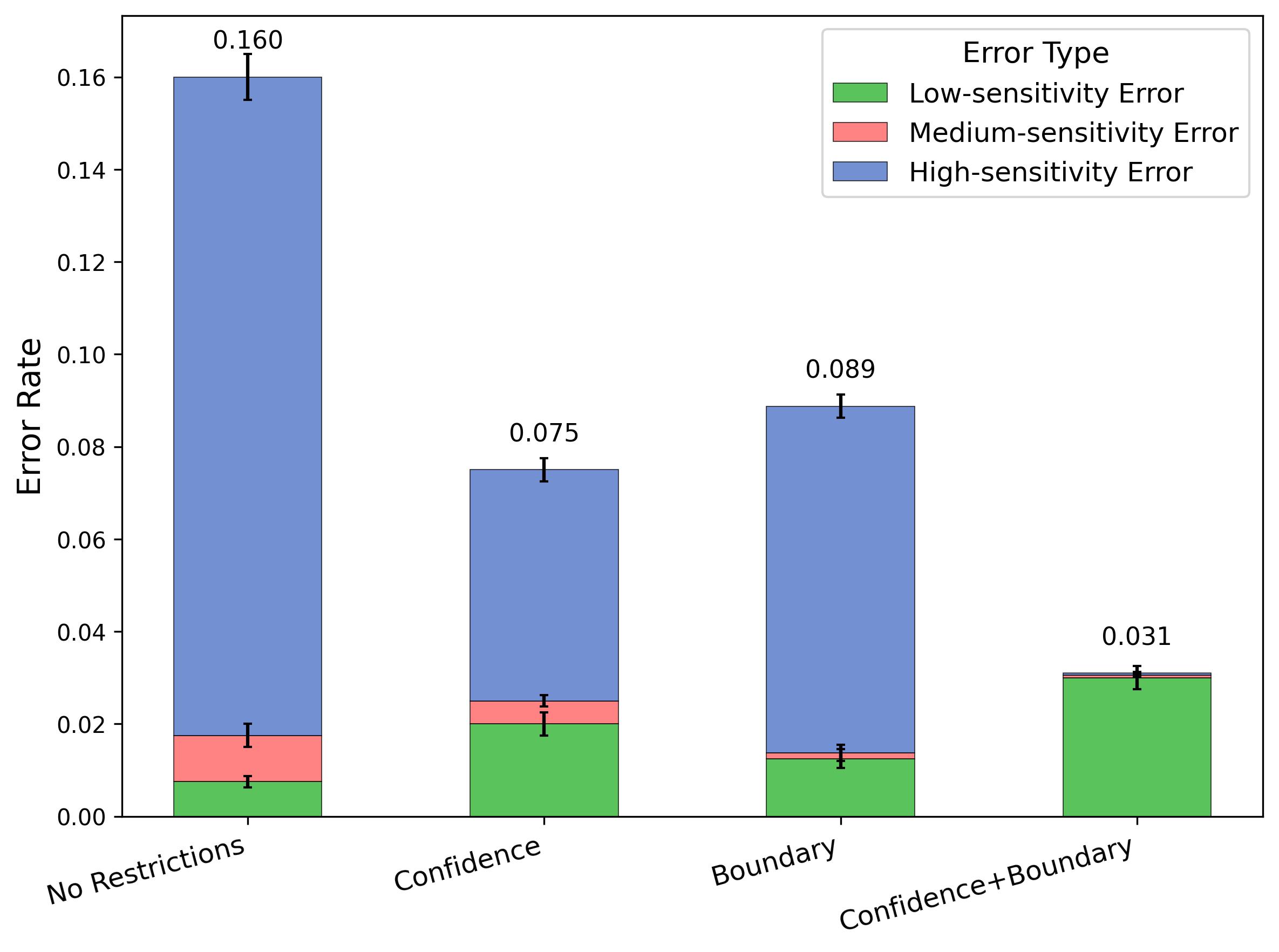}
      \caption{Action prediction error rate under different restriction strategies}\label{fig:error_rate_diff_restrictions}
      \vspace{-5pt}
   \end{figure}

The second strategy is the physical "boundary" restriction. The "boundary" is a specific range of working area where a human operates when doing specific actions. We consider both the current and predicted human trajectory to determine whether the human's hand is within the working area boundary. If the hand satisfied boundary restriction, we proceed with the action prediction model. Otherwise, we label the action as "no action."
The error rates for different sensitivities under various strategies are shown in \cref{fig:error_rate_diff_restrictions}. As demonstrated, the combination of "confidence + boundary" restriction effectively reduces the two types of most sensitive error rates.

These results demonstrate that the hierarchical human pose detection method and the sensitivity-aware method action prediction method used in our HRC framework significantly mitigate the environment disturbances and improve the action prediction accuracy. 

\subsubsection{H3 - Time Efficiency} \label{subsubsec:exp_h3}
Similar to \cref{subsubsec:exp_h1}, \Cref{tab:time consumption} compares the average completion times of six experimental groups across three sub-tasks, expressed as percentages relative to the longest recorded time for each task. As the results show, the group with multimodal perception and hierarchical plan prediction (''PP = 1, PM = 2") achieves the shortest completion time across all tasks, with an average reduction of 15.85\% compared to groups lacking both plan prediction and multimodal perception (''PP =0, PM = 0\&1"). Specifically, group with multimodal perception (''PM = 2") reduces task completion time by 8.54\% compared to single-modality groups (''PM = 0\&1"), while hierarchical plan prediction (''PP = 1") further decrease completion times by 7.42\% compared to groups without it (''PP = 0"). These results demonstrate that the proposed framework effectively combines the capabilities of multimodal perception and hierarchical plan prediction to optimize task efficiency, enabling seamless and proactive human-robot collaboration.
\begin{table}[ht]
    \centering
    \begin{tabular}{llccc}
    \hline
    \textbf{PP} & \textbf{PM} & \textbf{Task1 (\%)} & \textbf{Task2 (\%)} & \textbf{Task3 (\%)} \\
    \hline
    {0} & 0 & $78.80 \pm 13.57$ & $86.80 \pm 10.56$ & $53.19 \pm 18.31$  \\
    & 1 & $80.05 \pm 7.23$ & $79.34 \pm 12.49$ & $62.33 \pm 14.54$ \\
    & 2 & $75.86 \pm 15.67$ & $72.56 \pm 6.12$ & $45.12 \pm 13.29$ \\
    \hline
    {1} & 0 & $75.70 \pm 10.56$ & $80.46 \pm 13.98$ & $48.42 \pm 20.90$ \\
    & 1 & $72.02 \pm 9.10$ & $74.19 \pm 10.83$ & $43.72 \pm 13.50$ \\
    & 2 & $\textbf{67.42} \pm \textbf{4.91}$ & $\textbf{65.83} \pm \textbf{3.86}$ & $\textbf{39.45} \pm \textbf{14.83}$ \\
    \hline
    \end{tabular}
    \caption{The average completion time with standard deviation (only calculate success cases).}
    \label{tab:time consumption}
    \vspace{-5pt}
\end{table}

\subsubsection{H4 - Adaptability}
We employ an online adaptation strategy for the human physical action prediction model to refine its predictions.
We conducted experiments to evaluate the performance of an offline-trained prediction model across four different users. Accuracy was used to evaluate intention prediction, while mean-squared-error (MSE) was used to evaluate trajectory prediction.
\Cref{tab:adapt_res} presents the results of predictions with and without online adaptation. The results clearly show that online adaptation improves the performance of both action and trajectory prediction. The improvement in trajectory prediction is particularly significant, with a 24\% reduction in MSE. This is expected, as different individuals may have varying pose and pose preferences. The results demonstrate that the proposed framework effectively adapts to different human behaviors. 

\begin{table}[ht]
\centering
\begin{tabular}{l|c|c}
\hline
                   & w/o adaptation & with adaptation \\ \hline
Intention Accuracy & 0.9617         & \textbf{0.9626} \\ \hline
Pose MSE     & 1.2574         & \textbf{1.017}  \\ \hline
\end{tabular}
\caption{Prediction results with and without online adaptation.}
\label{tab:adapt_res}
\vspace{-10pt}
\end{table}

\subsubsection{H5 - User Satisfaction}

\begin{figure}[thpb]
      \centering
      \includegraphics[width=0.8\linewidth]{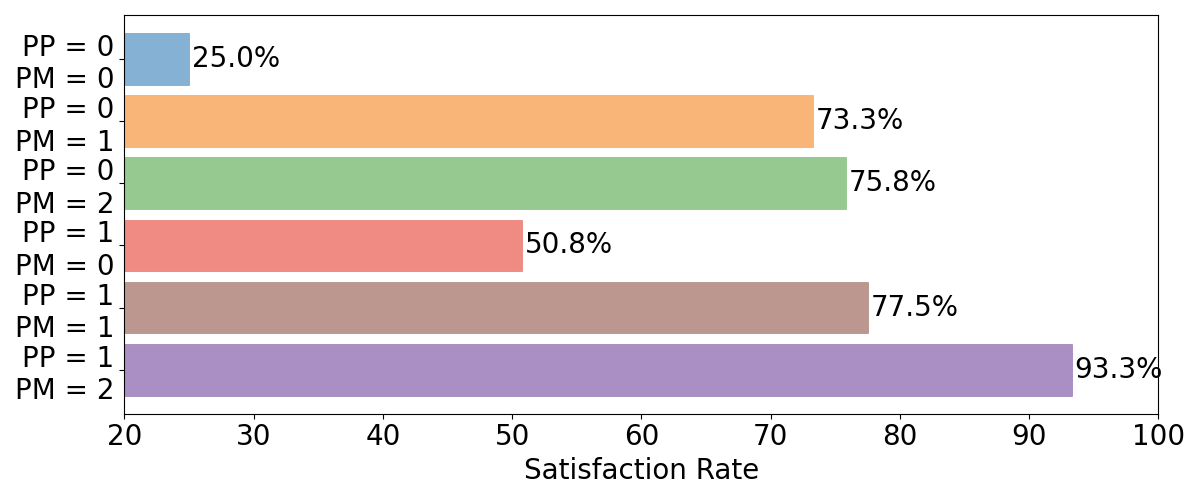}
      \caption{The user plan satisfaction rate, as calculated based on 120 trials across all three sub-tasks for each group by comparing the recorded user plan sequences with the actual execution sequences.}
      \label{fig:user_expectation}
      \vspace{-15pt}
   \end{figure}

\Cref{fig:user_expectation} shows the user plan satisfaction rate collected from six groups across three subtasks. As can be seen from the figure, with hierarchical plan prediction and multimodal perception (''PP = 1, PM = 2"), we can achieve the largest user plan satisfaction rate. Besides, with hierarchical plan prediction (''PP = 1"), we can improve the user plan satisfaction rate by 47.5\% than without hierarchical plan prediction (''PP = 0"). And with multimodal perception (''PM = 2"), we can improve the user plan satisfaction rate by 27.9\% than without multimodal perception (''PM = 0" and ''PM = 1"). These results prove that both the hierarchical plan prediction method and the multimodal perception can significantly improve the user plan satisfaction rate.

\begin{figure}[thpb]
      \centering
      \includegraphics[width=1.05\linewidth]{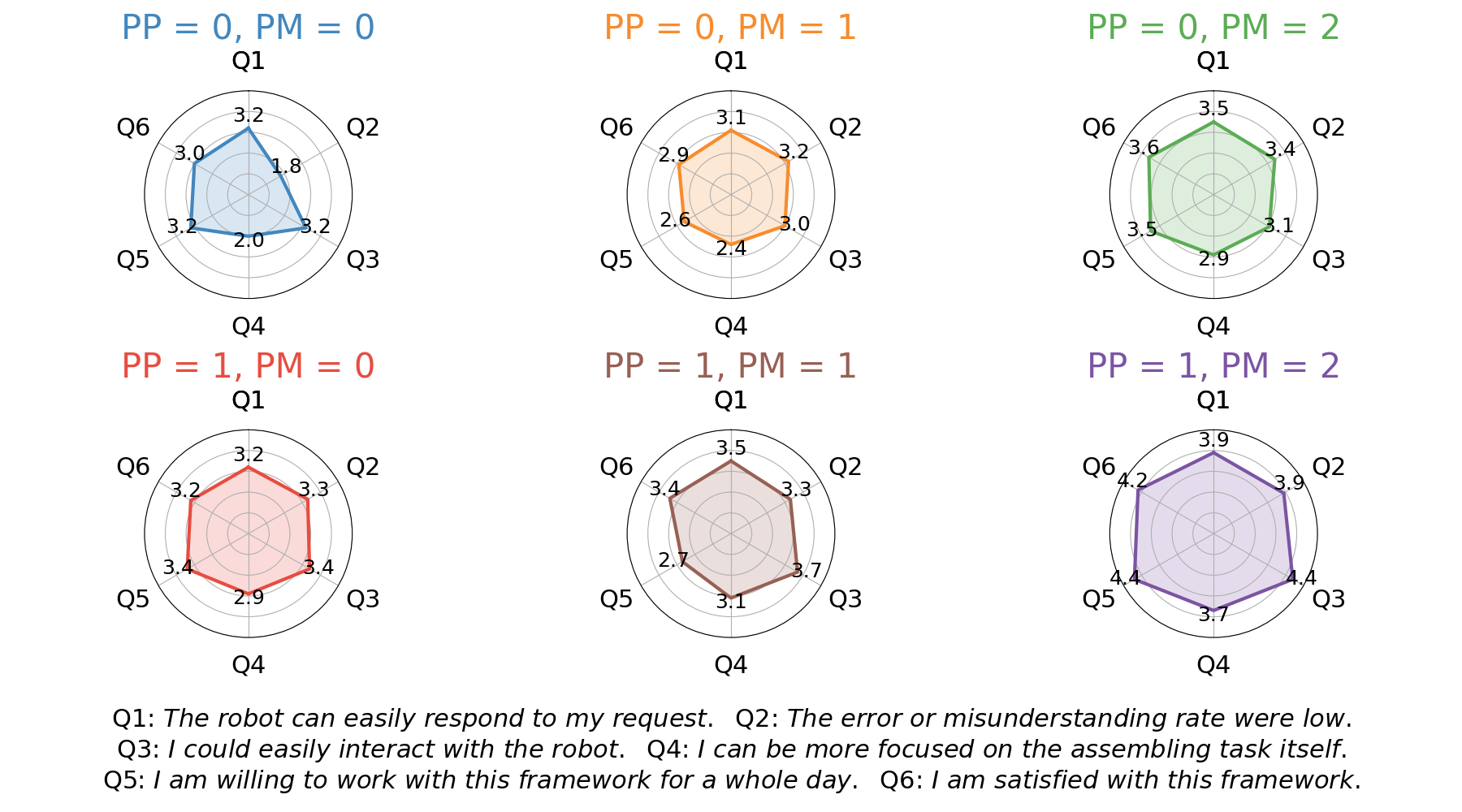}
      \caption{Quantified results of the user survey, with scores calculated as weighted sums of responses. Each question was rated on a five-point scale: ''Strongly Disagree: 1", ''Disagree: 2", ''Neutral: 3", ''Agree: 4", and ''Strongly Agree: 5".}
      \label{fig:process_user_study}
      \vspace{-10pt}
   \end{figure}
   
To gain deeper insights into the user experience with our framework, we conducted a survey featuring six carefully designed questions: timely responsiveness (Q1), misunderstanding rates (Q2), ease of interaction (Q3), support for maintaining focus on the primary task (Q4), willingness to work with the framework for extended periods  (Q5) and overall satisfaction (Q6). These questions evaluate the framework's user-friendliness and overall effectiveness, providing valuable insights into its ability to enhance user flexibility and satisfaction. 

\Cref{fig:process_user_study} shows quantized user satisfaction scores for six groups across six survey questions. Specifically, user feedback highlighted that: 1) Users found the framework highly responsive, accurate, and easy to interact with (Q1, Q2, Q3), reflecting an improved overall user experience. 2) Users reported greater focus on primary tasks rather than the collaboration process with the robot (Q4), indicating a higher user flexibility. 3) Users expressed a greater willingness to collaborate with the hierarchical and multimodal framework over an extended period (Q5), suggesting that it effectively reduces user workload. 4)  Overall, users reported higher satisfaction when working with the hierarchical and multimodal framework (Q6).
This user study demonstrates the effectiveness of the proposed hierarchical and multimodal framework in creating a user-friendly HRC system for long-term HRC tasks.

\section{LIMITATIONS}
While this work significantly showcases its great efficacy in accurate understanding of human-related behaviors, it does not address object-related understanding. However, an accurate understanding of real-time assemble structure would greatly contribute to better human plan prediction and task progree inference. Also, in this work, we assume no uncertainty regarding the location of objects. However, minimizing object-related uncertainties remains a substantial challenge for future work. Furthermore, future work could explore the use of powerful world models to model human behaviors for more flexible understanding of human actions and plans, though this may require a trade-off between flexibility and precision. Additionally, the task graph in this work requires manual design for specific scenarios; automating this process in future research would improve scalability to diverse tasks.
\section{CONCLUSIONS}
In this work, we identify the key challenges of long-term HRC tasks, present a structured task graph and propose a multimodal and hierarchical framework to address these challenges. We deploy our framework in a real-world long-term HRC assembly scenario and conducted extensive experiments and user studies. The result demonstrate that the proposed framework significantly improves the accuracy, robustness, efficiency, adaptability, and human satisfaction during the long-term HRC process, demonstrating the framework's great potential to further long-term HRC applications.





\bibliographystyle{plainnat}
\bibliography{references} 


\newpage
\clearpage
\onecolumn
\appendix
\subsection{Proof That the Multimodal Framework Yields Greater Mutual Information }\label{subsec:proof of entropy reduction}

In \cref{subsec:multimodality}, we demonstrated that the multimodal framework achieves greater mutual information between the goal and the observations compared to a single-modal framework. In this section, we provide a formal proof of this claim by establishing the following inequality:
\begin{align}
    I(g; O_V, O_A) \ge \max \{  I(g; O_{V}) ,  I(g; O_{A}) \}
\end{align}

The robot's performance in achieving the goal $g$ is enhanced by these informative observations. We consider the mutual information between the goal and observations:
\begin{align}
    I(g; O_V, O_A) =  H(g) - H(g|O_V, O_A) \label{eq:i_mm}
\end{align}
where $I(\cdot,\cdot)$ denotes mutual information and $H(\cdot)$ denotes entropy. 

The conditional entropy is defined as:
\begin{align*}
    H(g|O) &= H(g,O) - H(O) \\ 
    &= -\mathbb{E}[\log P(g|O=o)]
\end{align*}

This definition leads to the following rules:
\begin{align*}
    & H(g,O_V) = H(O_V) + H(g|O_V) \\
    & H(g,O_V,O_A) = H(O_V) + H(O_A|O_V) + H(G|O_A,O_V)
\end{align*}

From this, we can derive:
\begin{align*}
    & H(g|O_V) - H(g|O_V,O_A) \\
    &= H(g,O_V) - H(O_V) - H(g,O_V,O_A) + H(O_V) + H(O_A|O_V) \\
    &= H(g,O_V) - H(g,O_V,O_A) + H(O_A|O_V) \\
    &= H(g,O_V|O_A) + H(O_A|O_V) > 0
\end{align*}

Since both \(H(g,O_V|O_A)\) and \(H(O_A|O_V) \) are positive, we can conclude that \(H(g|O_V) >H(g|O_V,O_A)\). Similarly, we have \(H(g|O_A) >H(g|O_V,O_A)\).
This demonstrates that incorporating more modalities can reduce entropy and provide more information.
 Combining the above expressions, we find:
\begin{align}
    I(g; O_V, O_A) & \ge \max \{ H(g) - H(g|O_V) , H(g) -  H(g|O_A)\} \nonumber  \\
    & = \max \{  I(g; O_{V}) ,  I(g; O_{A})\} 
\end{align}
Therefore, we conclude that the multimodal framework yields greater mutual information between the goal and the observations than a single-modal framework, significantly enhancing the robot’s capability to handle complex and long-term collaboration tasks.

 \subsection{Sensitivity-aware Action Prediction} \label{sec:sens_int_pred}
 \begin{figure}[ht]
  \centering
  \subfigure[Naive prediction]{\includegraphics[scale=0.32]{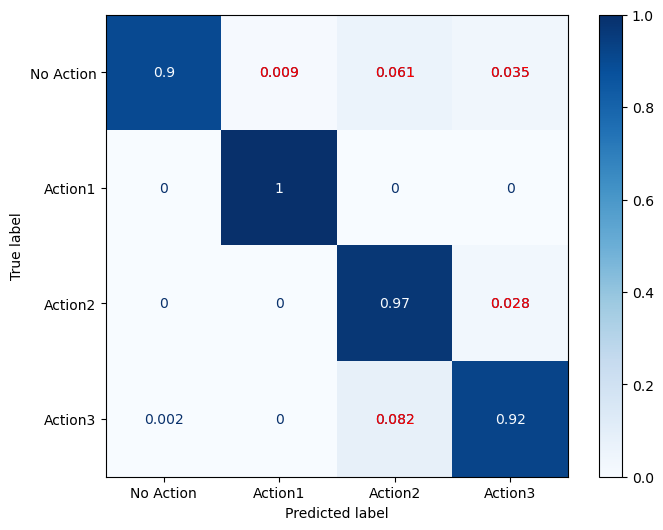}\label{fig:cm_no_restrict_test_self}}
  \subfigure[Confidence-based restriction]{\includegraphics[scale=0.32]{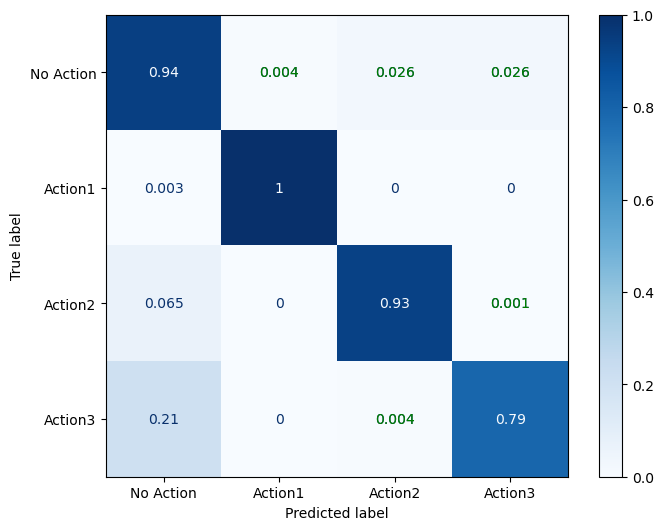}\label{fig:cm_ood_restrict_test_self}}
  \caption{Confusion matrix for intention prediction with and without confidence-based restriction}
  \label{fig:restriction_comparison}
\end{figure}

In our HRC framework, the robot's motion is highly dependent on the predicted human actions. However, not all prediction errors have the same consequences or remedies. For instance, if the human is actually performing a specific action $a_1$,  it might be acceptable to wrongly classify it as "no action" $a_0$. Because when the robot detects "no action," it will not move and since human actions is a sequence, it might be highly possible that at some point within this action sequence, the robot would detect the action appropriately. On the other hand, if the real action $a_1$ is misclassified as a different action $a_2$, the robot will generate motions based on $a_2$ and move to the wrong location, which could lead to safety issues. Similarly, misclassifying "no action" as another action could also result in critical safety concerns.
To address this, we assign different sensitivity scores to various types of misclassification errors and optimize the prediction model to minimize the most critical errors. The sensitivity scores assigned to each type of error are shown in \cref{tab:Sensitivity_score} of \cref{subsec:human_physical_action_prediction}.

To reduce more sensitive errors, we propose two different restriction strategies: confidence and boundary restriction, as presented in \cref{subsubsec:exp_h2}. \Cref{fig:restriction_comparison} shows the confusion matrix for action prediction with and without the confidence-based restriction. As seen in the figure, the error rate for high-sensitivity misclassifications, such as predicting "no action" as other actions, is significantly reduced with the confidence-based restriction strategy.

\subsection{Video Demo}
\begin{figure}[htbp]
    \centering
    \href{https://youtu.be/uQEDsGZp1uM?si=OvpvpHi14W3vtwQs}{%
        \includegraphics[width=0.6\linewidth]{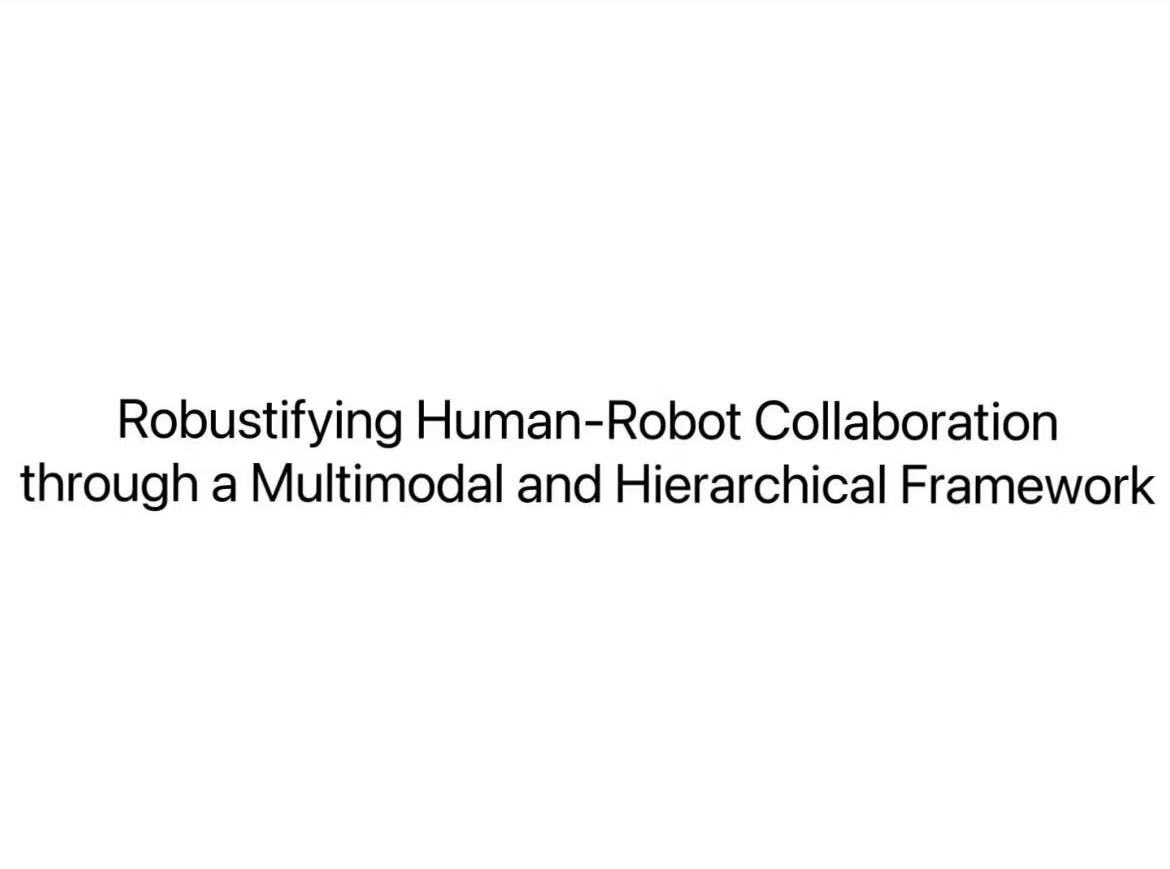}%
    }
    \caption{Car assembly task demonstration. Click the image to watch the video.}
    \label{fig:video_demo}
\end{figure}

The car assembly task, serving as an example of long-term HRC, is demonstrated in \cref{fig:video_demo}. Click the image to watch the video.
The results show that the proposed multimodal and hierarchical framework significantly improves the robustness, efficiency, and flexibility of HRC.

\end{document}